\documentclass[runningheads]{llncs}
\usepackage[T1]{fontenc}
\usepackage{bm}
\usepackage{xcolor}
\usepackage[bottom]{footmisc}
\usepackage{booktabs}
\usepackage{cite}
\usepackage{balance}
\usepackage{amsmath,amssymb,amsfonts}
\usepackage{algorithmic}
\usepackage{comment}
\usepackage{graphicx}
\usepackage{textcomp}
\usepackage{xcolor}
\usepackage{booktabs}
\usepackage{hyperref}
\usepackage[acronym]{glossaries}
\makeglossaries
\usepackage{cite}
\usepackage{balance}
\usepackage{multicol}
\usepackage{amsmath,amssymb,amsfonts}
\usepackage{algorithmic}
\usepackage{graphicx} 
\usepackage{hyperref} 
\usepackage[pdf]{graphviz}
\usepackage{booktabs}
\usepackage{subfig}
\usepackage{comment}

\usepackage{tikz}
\usepackage[table]{xcolor}
\usetikzlibrary{shapes.geometric}
\usetikzlibrary{positioning}
\usetikzlibrary{shapes,arrows,positioning}
\usepackage{amsmath}
\usepackage{textcomp}
\usepackage{xcolor}
\usepackage{hyperref}
\usepackage{graphicx}
\begin{document}
%
\title{MambaBEV: A BEV-based 3D detection model with Mamba2
\thanks{This work is partially supported by National Natural Science Foundation of China (NSFC) under Grant 52372410. All correspondence should be sent to J. Wang (Email: wangjx@seu.edu.cn).}
\thanks{This paper has been accepted by ICPR 2026.}}

%
%
\author{Zihan You\inst{1} \and
 Ni Wang\inst{2} \and
Hao Wang\inst{3} \and
 Qichao Zhao\inst{3} \and
 Jinxiang Wang\inst{4}}

%
\authorrunning{Z. You et al.}
%
\institute{School of Instrument Science and Engineering, Southeast University,   China \and
Amazon Development Center Germany GmbH, Germany \and
 T3CAIC Technology, China\and
 School of Mechanical Engineering, Southeast University, China }

%

\newacronym{bev}{BEV}{Bird's Eye View}
\newacronym{ssm}{SSM}{State-Space Model}

\maketitle              
%
\begin{abstract}

Accurate 3D object detection in autonomous driving relies on Bird’s Eye View (BEV) perception and effective temporal fusion. 
However, existing fusion strategies based on convolutional layers or deformable self-attention struggle to model global context in BEV space, leading to reduced accuracy for large objects.
To address this limitation, we propose MambaBEV, a novel BEV-based 3D object detection model that leverages Mamba2, an advanced state-space model (SSM) optimized for long-sequence processing. 
Our key contribution is TemporalMamba, a temporal fusion module that enhances global context modeling through a BEV feature discrete rearrangement mechanism tailored for sequential processing. 
In addition, we introduce a Mamba-based DETR head to improve multi-object representation. 
Evaluations on the nuScenes dataset demonstrate that MambaBEV-base achieves 51.7\% NDS and an 42.7\% mAP. 
Furthermore, evaluation within an end-to-end autonomous driving paradigm validates its effectiveness in motion forecasting and planning.
These results highlight the potential of state-space models for improving global context understanding and large-object detection in autonomous driving perception systems.
\end{abstract}

\section{INTRODUCTION}

Ensuring accurate and reliable 3D object detection is crucial for autonomous driving systems, directly impacting safety and path planning. 
Traditional perception methods, such as the Hough Transform\cite{kumar2022efficient} and keypoint-based feature extraction \cite{yuan2022keypoints}, laid the groundwork for object detection but struggled with limited robustness and scale variance. 
The rise of deep learning-based perception has significantly improved detection accuracy, yet challenges remain, particularly for monocular camera-based methods\cite{han2024monocular}, which suffer from depth estimation errors and blind spots, thereby posing risks to vehicle safety.

To tackle these issues, researchers have explored multi-camera perception systems, such as binocular stereo matching\cite{yang2023improved} and surround-view camera networks, which provide richer geometric cues than monocular setups.
While these approaches improve distance estimation, they may introduce challenges such as feature redundancy and difficulty in cross-camera target re-identification, especially when information from different views is not explicitly aligned.
A more promising solution is \gls{bev} -based 3D object detection, which consolidates multi-camera inputs into a unified top-down representation, enhancing distance estimation, obstacle detection, and cross-view information sharing\cite{huang2021bevdet}.

\begin{table}[htbp]  
\centering  
\vspace{-8mm}
\caption{average precision for bevformer-tiny\cite{li2203bevformer}}  
\label{average precision}
\begin{tabular}{lcccccccc}  
\toprule  
\textbf{categories} & \textbf{dist0.5$\uparrow$} & \textbf{dist1.0$\uparrow$} & \textbf{dist2.0$\uparrow$} & \textbf{dist4.0$\uparrow$} \\  
\midrule  
car & 0.0877 & 0.3366 & 0.6277 & 0.7809 \\  
pedestrian & 0.0251 & 0.1993 & 0.4604 & 0.6469 \\  
bicycle & 0.0056 & 0.1132 & 0.2902 & 0.3992 \\  
\midrule  
truck &0.0019& 0.0700 & 0.2587  & 0.4403 \\  
construction vehicle & 0.0000 &  0.0008& 0.0651 & 0.1696 \\  
bus & 0.0000 &  0.0588 & 0.3203 & 0.5619 \\  
\bottomrule  
\end{tabular}  
\label{average precision}
\vspace{-6mm}
\end{table}  

Another critical aspect of autonomous driving perception is the temporal aggregation. 
While single-frame detection provides a straightforward approach, it often suffers from occlusion, missed detections, and temporal inconsistency between frames. 
To address these limitations, temporal fusion techniques have been developed to incorporate historical features, significantly improving detection robustness and accuracy\cite{li2203bevformer}. 
Traditional temporal fusion methods, such as deformable self-attention mechanisms\cite{li2203bevformer}, dynamically sample spatial features and enhance computational efficiency compared to global self-attention.
However, these methods may struggle with global context modeling and long-range interactions due to their sparse and query-centric sampling strategies.
For example, on the COCO 2017 validation set, deformable attention-based models such as Deformable-DETR\cite{zhu2020deformable} reported a 2.9\% lower average precision (AP) on large objects compared to their global self-attention counterparts.

Similarly, in BEV-based 3D object detection, deformable self-attention models like BEVFormer\cite{li2203bevformer}  demonstrate higher accuracy for small objects (e.g., pedestrians, bicycles) but reduced performance for larger objects (e.g., trucks, buses) (Table~\ref{average precision}).  
This disparity can be partially attributed to the sparse sampling nature of deformable attention, which limits spatial coverage and lacks explicit mechanisms for global cross-scale interaction.
Even when increasing the number of sampling points, prior studies have shown that deformable attention primarily aggregates local features and may still fall short in capturing holistic spatial relationships.

\begin{figure*}[htbp]
\centerline{\includegraphics[width=1.0\textwidth]{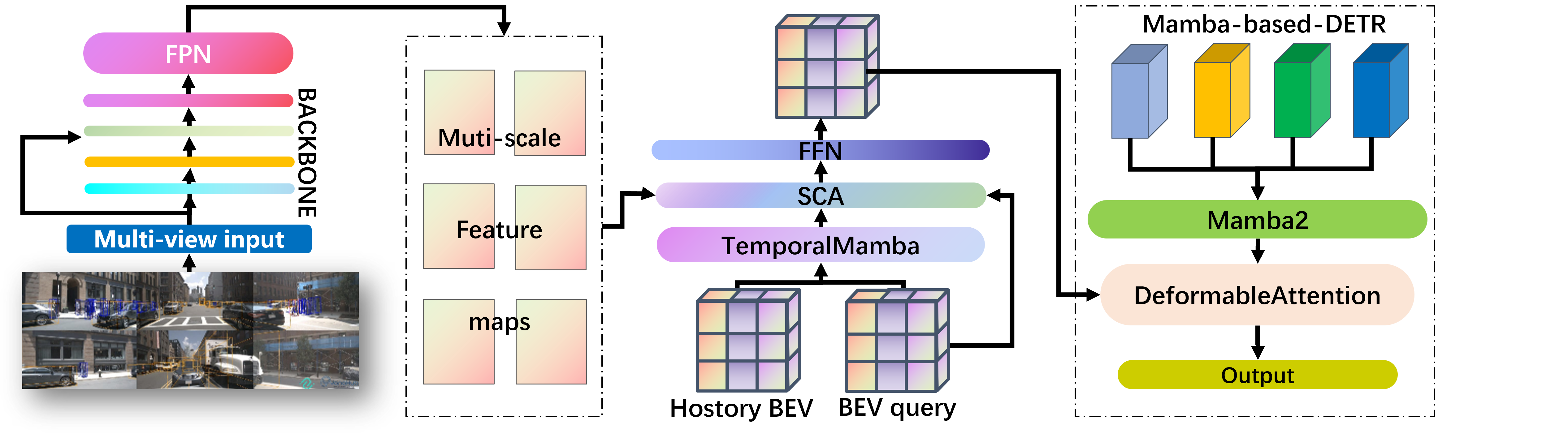}}
\caption{Given an RGB image captured by six surrounding cameras, a pretrained backbone generates six feature maps. These feature maps are processed through a Feature Pyramid Network (FPN) to extract multi-scale features. Subsequently, the Spatial Cross Attention (SCA) module performs backward projection to produce a BEV feature map. The TemporalMamba block then fuses historical BEV features with current BEV features, guiding the generation of new current BEV features. After several processing layers, a Mamba-based-DETR head serves as the 3D object detection head.}
\vspace{-6mm}
\label{overall}
\end{figure*}



Addressing these limitations is critical for enhancing 3D object detection performance in autonomous driving.
Recently, State-Space models (SSMs) have emerged as a compelling alternative to attention-based architectures for long-sequence modeling, as they explicitly model global dependencies through recurrent state transitions while maintaining linear computational complexity with respect to sequence length\cite{gu2023mamba}.
Unlike attention mechanisms that rely on sparse sampling or pairwise token interactions, SSMs propagate information through a global latent state, enabling holistic context aggregation without being constrained by local receptive fields. Among SSMs, Mamba introduces a selective state update mechanism that dynamically controls information flow based on input content, allowing it to capture long-range dependencies efficiently while remaining computationally scalable.
Mamba2 further refines this design by improving hardware efficiency and stability, making it more suitable for high-resolution and long-horizon sequence modeling tasks\cite{dao2024transformers}. These properties align well with the requirements of temporal BEV-based 3D perception, where features across large spatial extents and multiple time steps must be integrated globally and consistently.
Therefore, integrating Mamba2 into BEV-based 3D perception provides a principled alternative to deformable attention for temporal fusion, enabling explicit global temporal modeling with linear complexity and offering a strong inductive bias for large-object perception in autonomous driving.

Our contribution in the paper could be summarized as follows: 
\begin{itemize}
\item We introduce \textbf{MambaBEV}, a Mamba2-based 3D object detection model. To the best of our knowledge, this is the first work to integrate Mamba into a camera-based 3D detection framework.
\item We propose \textbf{TemporalMamba}, a temporal fusion module based on Mamba2 that models global temporal dependencies via state-space transitions rather than sparse attention mechanisms.
To bridge the gap between BEV feature topology and sequential state updates, we introduce a \textbf{BEV feature discrete rearrangement} strategy that enables effective sequence modeling over spatially structured BEV features.
\item We design a \textbf{Mamba-based DETR detection head}, replacing conventional attention-based decoding with a Mamba cross-sequence interaction mechanism, demonstrating the applicability of state-space modeling beyond temporal fusion.
\item We conduct extensive experiments on 3D object detection and end-to-end autonomous driving benchmarks following ~\cite{jiang2023vad}, including detailed ablations and analyses.
\end{itemize}

\section{Related work}

\subsection{camera-based 3D Object detection}

In the field of 3D object detection from images, several pioneering approaches have significantly advanced the domain. LSS\cite{philion2020lift} and BEVDet\cite{huang2021bevdet} represent notable methods that transform image-based features into a \gls{bev} representation, enabling more accurate 3D detection by leveraging the geometric properties of the scene.  Building upon this, BEVerse\cite{zhang2022beverse} enhances the BEV paradigm by integrating multi-view and temporal information, thereby improving the robustness and precision of 3D object detection.
BEVDepth \cite{li2023bevdepth} takes a step further by explicitly modeling depth information within the BEV framework. This addition allows for a more precise understanding of the spatial relationships between objects, leading to enhanced detection performance. 
DETR3D\cite{wang2022detr3d} introduces the transformer architecture to the 3D detection task, leveraging its powerful sequence-to-sequence modeling capabilities.
By generating a cost volume from a long history of image observations and augmenting per-frame monocular depth predictions with short-term, fine-grained matching, SOLOFusion\cite{park2022time} leverages both long-term and short-term temporal fusion to enhance object perception. VideoBEV\cite{han2024exploring} addresses the challenges of increasing computational and memory overheads associated with parallel fusion methods, demonstrating strong performance across various camera-based 3D perception tasks, including object detection, segmentation, tracking, and motion prediction. FB-BEV\cite{fbbev} and FB-OCC\cite{li2023fb} define the Spatial Cross Attention (SCA) process as backward projection, an inverse process of forward projection. They combine two processes into one method, which effectively improve the 3D detection capabilities of models.

\subsection{Mamba-based models}
Transformers have revolutionized various domains of deep learning with their ability to model sequences effectively\cite{vaswani2017attention}.
However, their application in long-sequence modeling remains computationally expensive due to quadratic complexity with respect to sequence length.
Addressing this, structured state space models (SSMs) have emerged as a scalable alternative, boasting linear complexity during training and constant state size during generation\cite{gu2023mamba}. 
Among the notable advancements in this field is the Mamba model, which offers a promising solution for efficiently handling long sequences without sacrificing performance.

The evolution of Mamba into Mamba2\cite{dao2024transformers} marked a significant leap, incorporating a structured state space duality (SSD) framework. 
This framework bridges SSMs with attention mechanisms, enhancing efficiency and scalability.
Mamba2 particularly excels in sequence processing tasks by employing block decompositions of semiseparable matrices, which optimize both computational and memory efficiency.
These innovations have set a new benchmark in sequence modeling, extending their application to complex tasks such as language processing and, crucially, to 3D object detection in autonomous driving systems.




\begin{figure*}[htbp]
\vspace{-6mm}
\centerline{\includegraphics[width=0.9\textwidth]{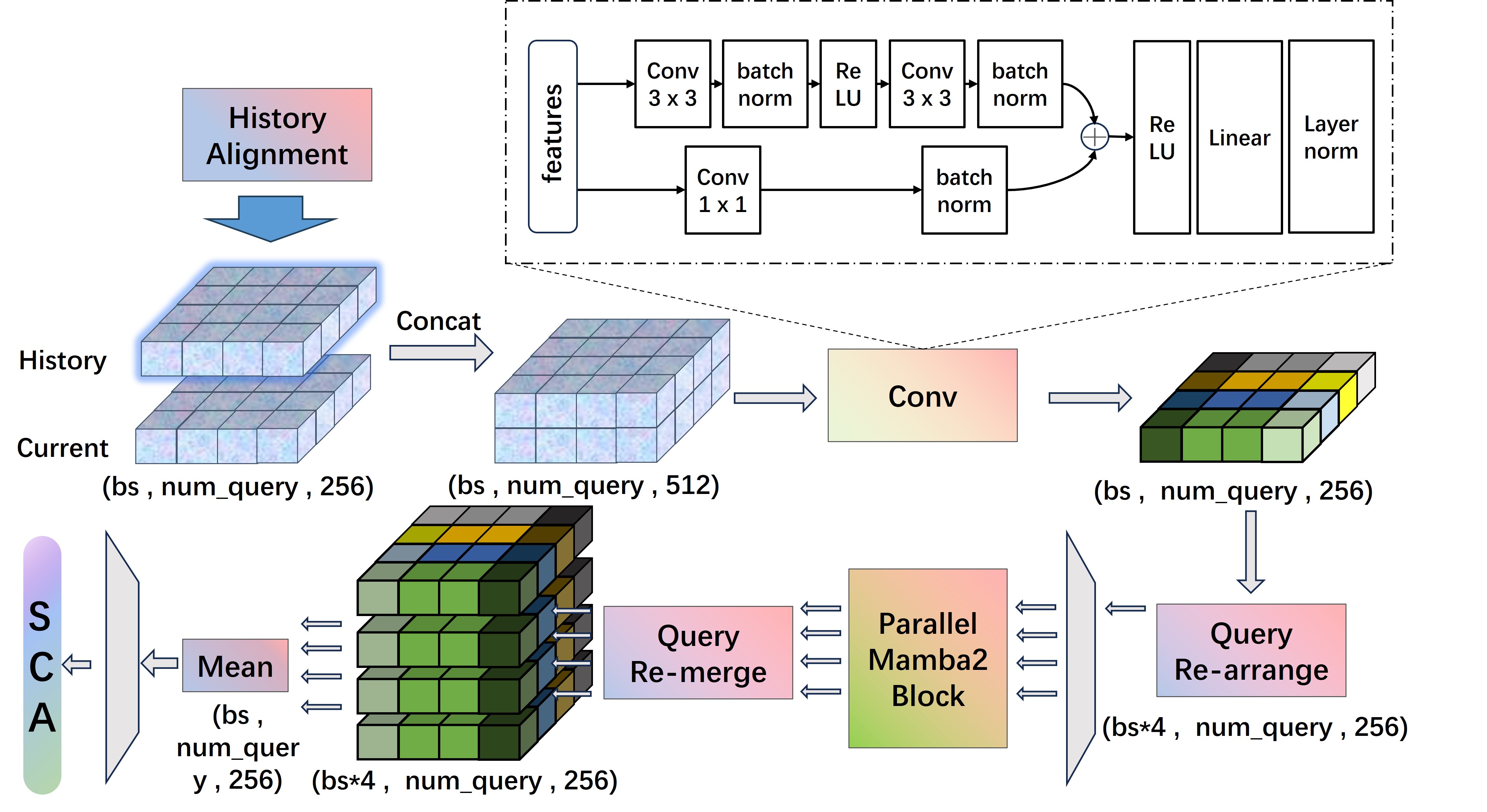}}
\caption{Proposed TemproalMamba module architecture with five parts Alignment, Compressed, Re-arrange, scan, Re-merge; 
Query Re-arrange and Query Re-merge is shown in Fig~\ref{Re-arrange} and Fig~\ref{fig3} }
\label{temporalmamba}
\vspace{-6mm}
\end{figure*}

Recent developments have expanded the capabilities of Mamba-based models. 
The introduction of Vision Mamba(Vim)\cite{zhu2024visionmambaefficientvisual} showcases a bidirectional state space model that enhances visual representation learning. Vim addresses the challenges of positional sensitivity and the need for global context in visual data, marking a significant advancement over traditional self-attention mechanisms. 
The VMamba\cite{liu2025vmamba} model transitions Mamba's capabilities into the vision domain with its Visual State-Space (VSS) blocks and the innovative 2D Selective Scan (SS2D) module, optimizing contextual information gathering.
Voxel Mamba\cite{zhang2024voxel} utilizes a group-free strategy to handle 3D voxel data, maintaining spatial proximity and enhancing detection accuracy without the computational overhead typical of Transformers. This model's application to point cloud data for 3D object detection exemplifies the potential of SSMs to revolutionize the processing of spatial data.

These advancements illustrate the evolving landscape of Mamba-based models, highlighting their potential to redefine the efficiency and efficacy of autonomous systems requiring complex sequence and spatial data processing.

\section{Methodology}
MambaBEV employs an advanced SSM comprising two primary components. The first is the TemporalMamba block, a fusion engine based on a proposed Mamba-CNN architecture, which integrates BEV features across sequential frames to enhance temporal consistency and detection robustness. The second component is the Mamba-based-DETR, an innovative decoder head that processes the fused features to accurately localize and classify 3D objects.

\vspace{-3mm}
\subsection{Architectural Design \& Feature Encoding}
\vspace{-1mm}
The MambaBEV system architecture, illustrated in Figure~\ref{overall}, integrates four essential components to process input from six RGB cameras. Initially, inputs are handled by the image feature encoder, which utilizes a robust backbone comprising ResNet-50 pretrained on ImageNet, ResNet-101-DCN initialized from the FCOS3D checkpoint, to efficiently extract high-level features from each image. The extracted features are then enhanced using a Feature Pyramid Network (FPN), generating multiscale features crucial for detecting objects at various scales.

These multi-scale feature maps are subsequently processed by the Spatial Cross Attention (SCA) module, producing a unified \gls{bev} feature map. The TemporalMamba module enriches this integration by fusing historical and current BEV features, thereby enhancing the feature context for accurate object detection. The enriched features undergo further refinement through several processing layers before the Mamba-based-DETR head analyzes them for final object detection.

\vspace{-3mm}
\subsection{TemporalMamba block}
\vspace{-1mm}

Traditional temporal fusion strategies for BEV-based 3D object detection rely on deformable self-attention, which dynamically samples spatial features to aggregate historical and current BEV features. 
The Temporal Self-Attention (TSA) module,  for instance, operates as follows: Given historical BEV feature maps and current feature maps, TSA concatenates them and uses a linear layer to generate attention weights and offsets. Each query, representing BEV features, is then calculated based on these weights in parallel.

\begin{figure}[htbp]
\centerline{\includegraphics[width=0.5\textwidth]{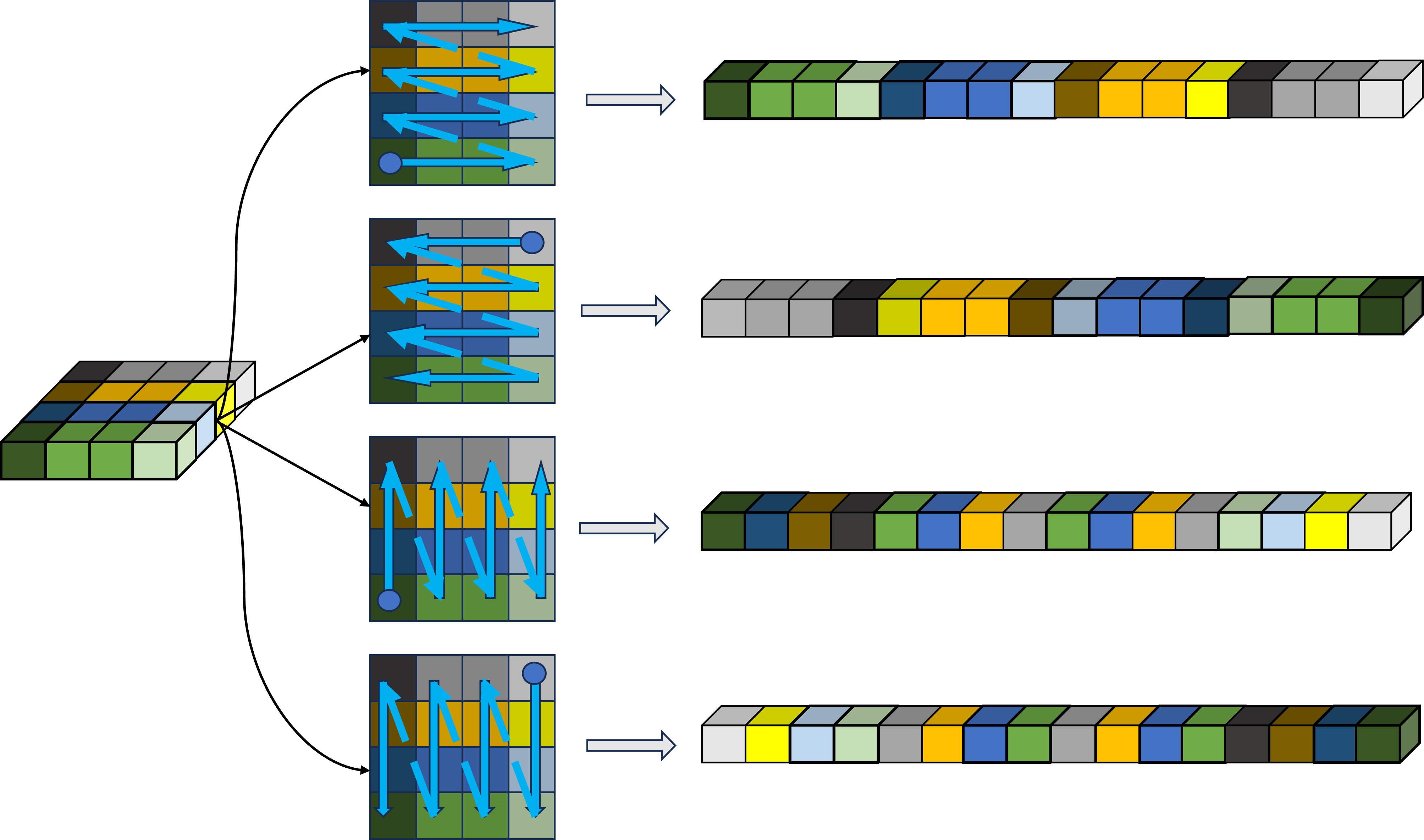}}
\caption{Query Re-arrange:  BEV feature map is discretely serialized and then recombined in four directions: forward-left, forward-upward, reverse-left, and reverse-upward. The recombined methods takes into account the impact of distance on the interaction of features, and adjust the methods in a balance way}
\label{Re-arrange}
\vspace{-6mm}
\end{figure}

Experiment results show that this paradigm has limitations. According to Table~\ref{average precision}, which lists some of the results of each categories, it is observed that these deformable self-attention module performs much better in detecting small objects like bicycle and pedestrian rather than large objects like bus and construction vehicle. Similar results are observed in other deformable attention-based models. The reason for this is that the mechanism does not facilitate cross-frame global interaction of large object features due to its restriction of allowing only three queries to interact with each reference query.

Our approach using Mamba2 increases global interaction capabilities. 
Initially, features from the previous frame are transformed using the ego rotation angle. As shown in Figure~\ref{temporalmamba}, given the transformed historical BEV feature map and the current feature map (both with dimensions of 256), we concatenate them along the third dimension. The concatenated features are then compressed from 512 to 256 dimensions using a convolutional block. This convolutional block consists of two parallel sub-modules: a 3×3 down-sampling convolutional layer that preserves important features while reducing dimensionality, and a pointwise convolutional layer that lowers the dimensionality and introduces non-linearity to learn complex patterns. To mitigate internal covariate shift, batch normalization is applied after each convolutional layer. The outputs of both sub-modules are concatenated and apply non-linear activation functions followed by linear layer and layernorm.

Next, we discretely rearrange Z and process it through the Mamba2 block. The Mamba2 block, originally designed for natural language processing and sequence processing, faces significant challenges when applied to vision-like data. Therefore, designing an appropriate discrete rearrangement method is crucial. We propose a four-direction rearrangement method, based on experimental results and inspired by Vmamba\cite{liu2025vmamba}. The impact of different rearrangement methods is discussed in the ablation study.

\begin{figure}[htbp]
\centerline{\includegraphics[width=0.5\textwidth]{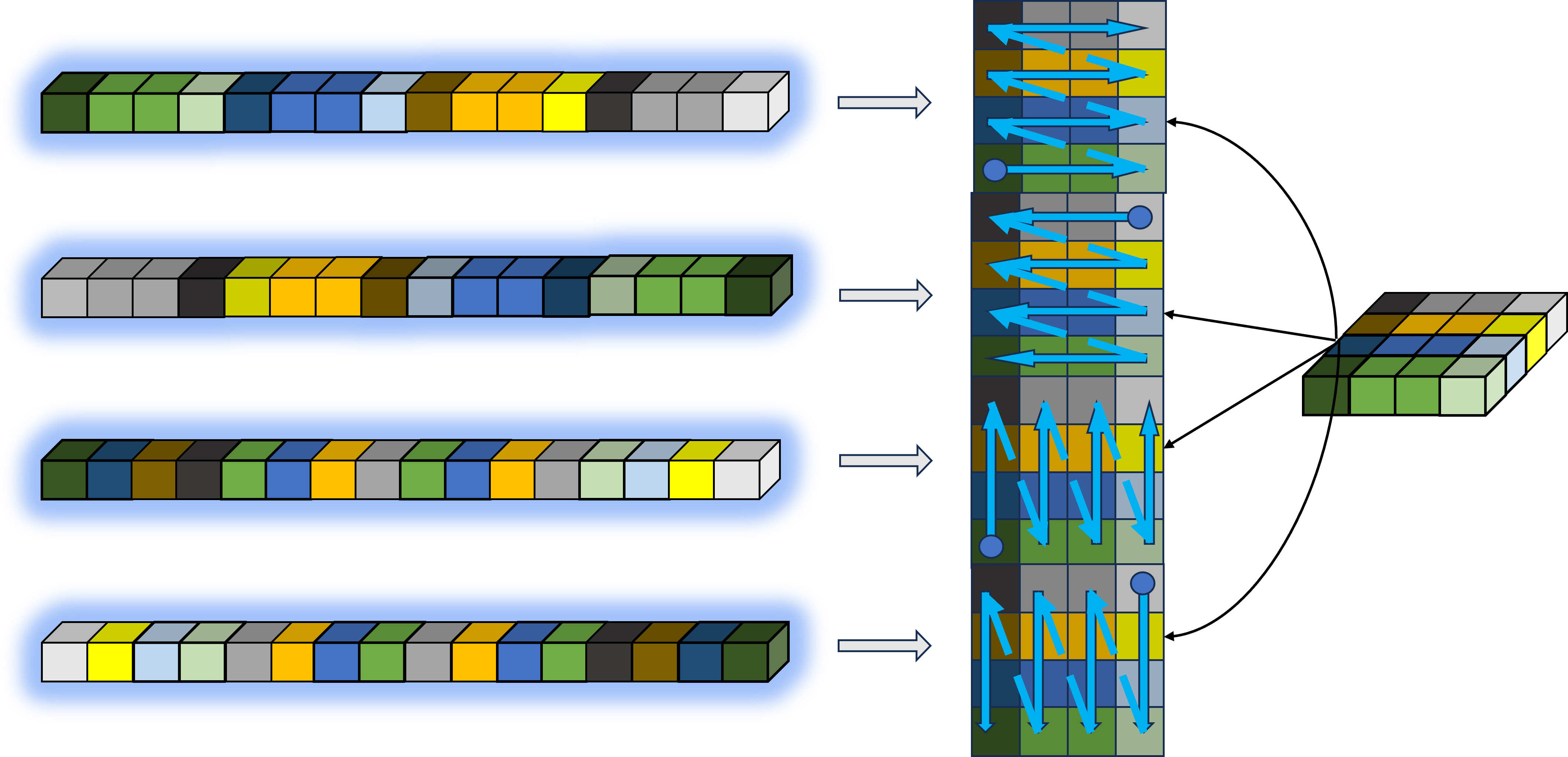}}
\caption{Query Re-merge:  
In this approach, we introduce a process termed "Query Re-merge," which serves as the inverse of the "Query Re-arrange" operation. Given four enhanced sequences, we initially segment them at the positions determined by the re-arrange operation. Subsequently, these segments are reassembled following the original partitioning scheme. To restore the sequences to their original dimensions, we apply an average calculation along the third dimension, resulting in a tensor of shape (batch size, number of queries, 256).}
\label{fig3}
\vspace{-5mm}
\end{figure}

A multi-directional feature sequence scanning mechanism is innovatively proposed, where the feature map Z is discretely serialized and then recombined in four directions: forward-left, forward-upward, reverse-left, and reverse-upward, as shown in Figure~\ref{Re-arrange}.  It is important to note that we do not adopt a serpentine, snake-like recombination approach, as we believe this results in an imbalance in the interaction between adjacent features, where some adjacent features may be close together while others are far apart. After Query Re-arrangement, the sequence is passed through a linear layer to project to a different dimensional space that fit mamba2 block. This new sequence is then fed into the Mamba2 model. The Mamba2 model outputs the enhanced sequence features, which incorporate and interact with global features. This helps to increase global awareness of the BEV space and aggregate cross frame features. The sequence is then recombined and restored to the original order as shown in Figure~ \ref{fig3}. We calculate the average of the four tensors, and the enhanced fused BEV feature map is added to the current BEV feature map with a dropout rate of 0.9 as a skip connection to avoid overfitting and reducing co-adaptation among features.

\subsection{Mamba-based-DETR head}

As depicted in Figure~\ref{overall}, we have redesigned the DETR head by integrating the Mamba2 architecture with the traditional DETR encoder and it is named mamba-based-DETR. Initially, 900 object queries undergo preprocessing and interact within the Mamba2 block, which functions similarly to self-attention mechanisms. The outputs from the Mamba block are then processed using deformable attention, akin to the traditional DeformableAttention.
\vspace{-3mm}
\section{Experiment}
\vspace{-3mm}
\subsection{Datasets}
\vspace{-2mm}
We evaluate our method on the nuScenes dataset \cite{caesar2020nuscenes} for 3D object detection and end-to-end planning.
The nuScenes dataset is a large-scale autonomous driving benchmark comprising 1,000 driving scenes collected across multiple cities, with approximately 280,000 annotated frames and comprehensive 3D object annotations. Model performance is assessed using the official nuScenes Detection Score (NDS), which jointly considers detection accuracy and multiple error terms. Specifically, NDS aggregates mean Average Precision (mAP) with several true positive metrics, including translation, scale, orientation, velocity, and attribute errors. This holistic metric ensures that both localization accuracy and motion-related factors are properly evaluated, providing a robust assessment of 3D detection performance. For end-to-end planning evaluation, displacement error and collision rate are adopted as the primary metrics to validate planning effectiveness and safety.

\subsection{Experimental settings}

We implement two variants of MambaBEV: MambaBEV-tiny and MambaBEV-base.
MambaBEV-tiny adopts an ImageNet-pretrained ResNet50 backbone, with a BEV grid of 50×50 at 2.048 m resolution, covering a perception range of 51.2 m in both X and Y directions. It uses input images of size 800×450, three historical frames, and a three-layer BEV encoder.
MambaBEV-base employs a ResNet101-CDN backbone initialized from the FCOS3D checkpoint (with V-99 as an alternative). It uses a denser 200×200 BEV grid at 0.512 m resolution, input images of 900×1600, four historical frames, and a six-layer BEV encoder, following DETR3D. Additionally, we evaluate MambaBEV in an end-to-end autonomous driving model, where it serves as the BEV feature backbone. This model is trained for 60 epochs on eight A800 GPUs with a learning rate of 1e-4, using a ResNet50 backbone, 1280×720 input images, 2000 map queries (100×20), and 300 agent queries. Other training settings follow MambaBEV-base.

\subsection{Main results}

\textbf{3D objects detection task} We evaluate our model on the nuScenes validation set for the 3D object detection task. As shown in Table~\ref{mainresult}, our MambaBEV-base outperforms BEVFormer-S by 3.51\% in mean Average Precision (mAP) and 5.97\% in the nuScenes Detection Score (NDS). The BEVFormer-S uses the Spatial Cross Attention (SCA) as the backward projection method, similar to our approach, but processes a single frame without any temporal fusion technic. This improvement highlights the effectiveness of our TemporalMamba block. Furthermore, the average velocity error decreases by 37\% when the TemporalMamba block is added, demonstrating that incorporating historical information, particularly processed by the TemporalMamba block, significantly enhances velocity estimation by providing historical data.

Compared to other methods leveraging temporal information, the Mamba block shows superior performance. For instance, our method achieves a 4.51\% improvement in mAP and a 6.37\% improvement in NDS over PolarDETR. Additionally, our method exhibits the lowest mean Absolute Velocity Error (mAVE) of 0.432, further validating its exceptional performance in velocity estimation. Our MambaBEV-base achieves a lower mean Average Translation Error (mATE) compared to the BEVFormer-s (0.669 vs 0.725), indicating that the predicted 3D bounding boxes are more accurately localized in space. Similarly, the mean Average Scale Error (mASE) decreases from 0.272 to 0.265, showing that object dimensions are better estimated.
\begin{table*}[htbp]  
\vspace{-6mm}
\centering  
\caption{Open loop planning performance}  
\label{end2endresult}
\begin{tabular}{lcccccccc}  
\toprule  
\textbf{Method} & \textbf{1s(L2)$\downarrow$} & \textbf{2s(L2)$\downarrow$} & \textbf{3s(L2)$\downarrow$} & \textbf{Avg$\downarrow$} & \textbf{1s(Col)$\downarrow$} & \textbf{2s(Col)$\downarrow$} & \textbf{3s(Col)$\downarrow$} & \textbf{Avg$\downarrow$} \\  
\midrule  
NMP &- & -&2.31   & - & - & - &1.92 & - \\  
ST-P3 & 1.33 & 2.11 & 2.90 & 2.11 & 0.23 & 0.62 & 1.27 & 0.71 \\  
ours & 1.03 & 1.76 & 2.53 & 1.77 & 0.25 & 0.84 & 1.72 & 0.93 \\  
\bottomrule  
\end{tabular}  
\end{table*}  

\begin{table}[htbp]  
\centering  
\vspace{-6mm}
\caption{Average precision of large objects from tiny-version}  
\label{average_precision_for_large_objects}  
\begin{tabular}{lcccccccc}  
\toprule  
\textbf{Methods}  & \textbf{categories} & \textbf{dist0.5$\uparrow$} & \textbf{dist1.0$\uparrow$} & \textbf{dist2.0$\uparrow$} & \textbf{dist4.0$\uparrow$}\\
\midrule  
 deformable & truck(\%) & 0.19 & 7.00 & 25.87 & 44.03 \\  
 ours & truck(\%) & 0.45  & 7.73 & 26.97 & 46.61 \\  
 \midrule  
 deformable & bus(\%) & 0.0 & 5.8 & 32.03 & 56.19 \\  
 ours & bus(\%) & 0.5  & 7.27 & 37.05 & 60.95 \\  
\bottomrule  
\vspace{-6mm}
\end{tabular}  
\end{table}

\begin{table*}[htbp]  
\centering
\vspace{1em}
\vspace{-8mm}
\caption{3D object detection results on nuScenes val. set}  
\label{mainresult} 
\begin{tabular}{lcccccccccc}  
\toprule  
\textbf{method}  & \textbf{imagesize} & \textbf{NDS$\uparrow$} & \textbf{mAP$\uparrow$} & \textbf{mATE$\downarrow$} & \textbf{mASE$\downarrow$} & \textbf{mAOE$\downarrow$} & \textbf{mAVE$\downarrow$} & \textbf{mAAE$\downarrow$} \\ 
\midrule  
BEVFormer-T  & 450*800 & 0.354 & 0.252  & 0.900 & 0.294 & 0.655 & 0.657 & 0.216 \\  
MambaBEV-T & 450*800 & \textbf{0.368} & \textbf{0.262}  & 0.881 & 0.2909 & 0.5998 & 0.6366 & 0.2209 \\
\midrule  
FCOS3D\cite{wang2021fcos3d}  & 1600*900 & 0.372 & 0.295 & 0.806 & 0.268 & 0.511 & 1.315 & 0.170 \\  
PGD\cite{wang2022probabilistic}  & 1600*900 & 0.409 & 0.335 & 0.732 & 0.263 & 0.423 & 1.285 & 0.172 \\  
DETR3D & 1600*900 & 0.425 & 0.346 & 0.773 & 0.268 & 0.383 & 0.842 & 0.216 \\  
BEVDet & 1056*384 & 0.384 & 0.317 & 0.704 & 0.273 & 0.531 & 0.940 & 0.250 \\  
PolarDETR\cite{chen2022polar}  & 1600*900 & 0.444 & 0.365 & 0.742 & 0.269 & 0.350 & 0.829 & 0.197 \\  
PolarDETR-T  & 1600*900 & 0.488 & 0.383 & 0.707 & 0.269 & 0.344 & 0.518 & 0.196 \\  
PETR\cite{liu2022petr}  & 1600*900 & 0.442 & 0.370 & 0.711 & 0.267 & 0.383 & 0.865 & 0.201 \\  
UVTR\cite{li2022unifying} & 1600*900 & 0.483 & 0.379 & 0.731 & 0.267 & 0.350 & 0.510 & 0.200 \\  
EPro-PnP-Detv2\cite{chen2022epro} & - & 0.490 & 0.423 & 0.547 & 0.236 & 0.302 & 1.071 & 0.123 \\  
DD3D\cite{park2021pseudo} & - & 0.477 & 0.418 & 0.572 & 0.249 & 0.368 & 1.014 & 0.124 \\  
BEVFormer-s  & - & 0.448 & 0.375 & 0.725 & 0.272 & 0.391 & 0.802 & 0.200 \\  
Ego3RT\cite{lu2022learning}  & - & 0.473 & 0.425 & 0.550 & 0.264 & 0.433 & 1.014 & 0.145 \\  
TempBEV\cite{monninger2024tempbev}& - & 0.508 & 0.408 & - & - & - & - & - \\  

MambaBEV-B  & 1600*900 & \textbf{0.517} & \textbf{0.427} & 0.669 & 0.265 & 0.365 & 0.468 & 0.193 \\  
\bottomrule  
\vspace{-12mm}
\end{tabular}  
\end{table*}

Table~\ref{average_precision_for_large_objects} presents the average precision for large objects, comparing deformable attention-based methods with our approach. We find that our model performs better in detecting large objects, such as trucks and buses, demonstrating its ability to facilitate global feature interaction and improve global context awareness within the bird’s-eye-view space.

Additionally, we observe that increasing the resolution of the bird’s-eye-view grid and incorporating additional frames significantly improve the model's performance. This is evidenced by the substantial performance gains in the base version compared to the tiny version, as more research is shown in the Ablation Study.

\begin{table}[htbp]  
\centering  
\vspace{-8mm}
\caption{Motion forecasting}  
\label{end2end_result_motion}  
\begin{tabular}{lcccccccc}  
\toprule  
\textbf{Method} & \textbf{minADE$\downarrow$} & \textbf{minFDE$\downarrow$} & \textbf{MR$\downarrow$} & \textbf{EPA$\uparrow$} \\  
\midrule  
PnPNet & 1.15 & 1.95 & 0.226 & 0.222 \\  
ViP3D & 2.05 & 2.84 & 0.246 & 0.226 \\  
Traditional & 2.06 & 3.02 & 0.277 & 0.209 \\  
Constant Pos. & 5.80 & 10.27 & 0.347 & - \\  
Constant Vel. & 2.13 & 4.01 & 0.318 & - \\  
Ours & \textbf{0.84} & \textbf{1.203} & \textbf{0.1478} & \textbf{0.546} \\  
\bottomrule  
\vspace{-8mm}
\end{tabular}  
\end{table}  

\textbf{End to end autonomous driving paradigms} Furthermore, we test our backbone in an end-to-end autonomous driving paradigm, with results presented in Table~\ref{end2endresult} and Table~\ref{end2end_result_motion}. The paradigms using our method demonstrate strong performance in open-loop evaluation on the nuScenes dataset. We assess the model in two key aspects: motion forecasting and planning. The performance is validated using L2 error and collision rate metrics. The results indicate that the end-to-end autonomous driving paradigm utilizing our method outperforms LiDAR-based methods in some instances. Additionally, we discuss the performance of motion forecasting in Table~\ref{end2end_result_motion} .

\vspace{-4mm}

\section{Ablation study}
\vspace{-2mm}
\subsection{Design of conv block}

To further analyze the effectiveness of the basic convolutional block, a fair comparison is made between the method that uses concatenation and the one that uses the convolutional methods within the TemporalMamba block. All methods in this study employ four-direction discrete rearrangement techniques. The concatenation method refers to the simple concatenation of historical BEV features with current BEV features along the vector dimension. As illustrated in Table~\ref{concatenates_convolutions}, the use of a convolutional block to fuse historical and current BEV features effectively improves model performance by approximately 1\%. Furthermore, it is evident that the convolutional block reduces the average velocity error, which is a crucial metric in the historical fusion module.

\begin{table}[htbp]  
\centering  
\vspace{-6mm}
\caption{Comparison between concatenates methods and Convolutions methods}  
\label{concatenates_convolutions}  
\begin{tabular}{lcccccccc}  
\toprule  
\textbf{Ways} & \textbf{backbone} & \textbf{NDS$\uparrow$} & \textbf{mAP$\uparrow$} & \textbf{mATE$\downarrow$}   & \textbf{mAVE$\downarrow$} & \textbf{mAAE$\downarrow$} \\  
\midrule  
Concat & R101 & 0.4936 & 0.399 & 0.696  & 0.513 & 0.213 \\  
Conv & R101 & 0.508 & 0.410 & 0.688  & 0.432 & 0.203 \\  
\midrule  
Concat & R50 & 0.3425 & 0.2518 & 0.9035  & 0.7612 & 0.2268 \\  
Conv & R50 & 0.3682 & 0.2622 & 0.8810  & 0.6366 & 0.2209 \\  
\bottomrule  
\vspace{-8mm}
\end{tabular}  
\end{table}

The improvement observed can be attributed to several factors. One key reason is the distinct types of features present in typical driving scenarios: dynamic (moving) features and static features. Concatenation-based fusion methods may fail to appropriately account for the interaction between these two feature types. Specifically, ego-motion and moving features can create discrepancies when fused without proper consideration of their spatial context. By employing a suitable receptive field, these issues can be mitigated, ensuring that related features are processed within the same receptive field.

\begin{table}[htbp]  
\centering  
\vspace{-8mm}
\caption{Impact cause by different window size}  
\label{different_window_size}  
\begin{tabular}{lcccccccc}  
\toprule  
\textbf{Window size} & \textbf{backbone} & \textbf{NDS$\uparrow$} & \textbf{mAP$\uparrow$} & \textbf{mATE$\downarrow$} & \textbf{mAVE$\downarrow$}  \\  
\midrule  
5 & R50 & 0.3642 & 0.2514 & 0.8844  & 0.6480 \\  
3 & R50 & 0.3682 & 0.2622 & 0.8810  & 0.6366 \\  
\bottomrule  
\vspace{-8mm}
\end{tabular}  
\end{table}


As shown in Table~\ref{different_window_size}, experiments indicate that a kernel size of 3 is optimal for feature fusion. Testing with different window sizes reveals that increasing the kernel size to 5 led to a decline in performance metrics. This suggests that overly large receptive fields may introduce noise, particularly when handling the distinct dynamic and static features found in typical driving environments.

\subsection{Impact of different discrete rearrangement methods}

 Mamba2 is specifically designed for natural language processing. However, for 3D object detection, the BEV feature map must be flattened or discretized before it can be processed by the Mamba block. Therefore, selecting an optimal discretization strategy is crucial. Motivated by \cite{liu2025vmamba}, two discretization methods are proposed: a one-direction discretization and a four-direction discretization approach.

\begin{table}[htbp]  
\centering
\vspace{-6mm}
\caption{Comparison between different rearrangement methods}  
\label{different_rearrangement}  
\begin{tabular}{lcccccccc}  
\toprule  
\textbf{Ways} & \textbf{backbone} & \textbf{NDS$\uparrow$} & \textbf{mAP$\uparrow$} & \textbf{mATE$\downarrow$}  & \textbf{mAOE$\downarrow$} & \textbf{mAVE$\downarrow$}  \\  
\midrule  
Single  & R50 & 0.3323 & 0.2390 & 0.9100  & 0.6596 & 0.8040 \\  
Four    & R50 & 0.3425 & 0.2518 & 0.9035  & 0.6545 & 0.7612  \\  
\bottomrule  
\vspace{-6mm}
\end{tabular}  
\end{table}

As demonstrated in Table~\ref{different_rearrangement}, the four-direction discretization method improves model performance. In the one-direction approach, the BEV queries (50×50) are discretized into a long sequence. While all methods have positional embeddings and implicitly encoded CAN bus information, the four-direction discretization method outperforms the one-direction method by 1.02\% in NDS and 1.28\% in mAP. This result supports the hypothesis that grouping more related queries closely enhances the Mamba block's ability to capture inter-query relationships. However, increasing the number of directions in the discretization method leads to higher parameter counts and computational complexity. Experiment shows that our methods provide an optimal solution. In this experiment, all models use the concatenation method.

\subsection{Impact of improving resolution of BEV features}

 \begin{figure}[hb]
 \vspace{-7mm}
\centerline{\includegraphics[width=0.6\textwidth]{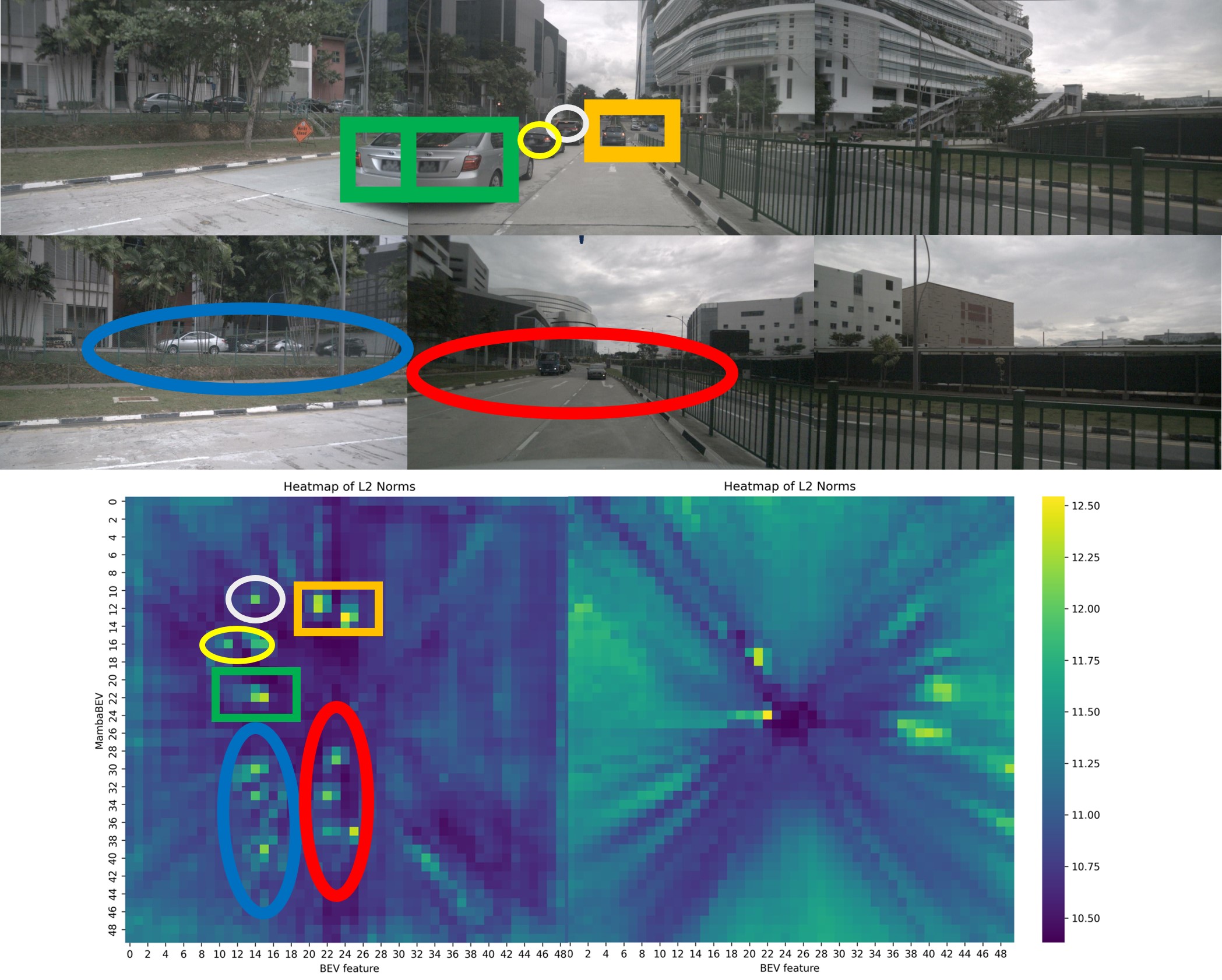}}
\vspace{-4mm}
\caption{visualization of BEV features}
\vspace{-8mm}
\label{fig4}
\end{figure}

To investigate the impact of resolution on model performance, the resolution of the MambaBEV-tiny BEV features is increased from 50 to 100. As shown in Table~\ref{resolution}, improving the resolution significantly enhances the model's performance across all metrics. In this experiment, convolutional block is employed instead of the concatenation strategy.

\begin{table}[htbp]
\vspace{-6mm}
\centering  
\caption{Impact of different resolution of BEV features}  
\label{resolution}  
\begin{tabular}{lcccccccc}  
\toprule  
\textbf{Resolution}  & \textbf{NDS$\uparrow$} & \textbf{mAP$\uparrow$} & \textbf{mATE$\downarrow$}   & \textbf{mAVE$\downarrow$} & \textbf{mAAE$\downarrow$} \\  
\midrule  
$100 \times 100$  & 0.4180 & 0.2951 & 0.7837  & 0.4732 & 0.1982 \\  
$50 \times 50$  & 0.3425 & 0.2518 & 0.9035 & 0.7612 & 0.2268 \\  
\bottomrule  
\vspace{-10mm}
\end{tabular}  
\end{table}  
\subsection{Visualization of BEV features}

 To further analyze the effectiveness of the TemporalMamba block, BEV features are visualized after passing through the TemporalMamba block. L2 norm is used to generate heatmaps for comparison. As illustrated in Figure~\ref{fig4}, it is evident that the model learns the object features effectively. 
Additionally, a comparative analysis is conducted between the TemporalMamba block and the deformable attention module. The left side of the figure illustrates the \gls{bev} features generated by the Mamba-based module, while the right side displays those produced by the deformable-based module. It is evident that the BEV features from the Mamba-based module exhibit superior global awareness, which intuitively accounts for its enhanced performance in detecting large objects. 
 

\vspace{-3mm}
\section{CONCLUSIONS}
\vspace{-3mm}
This paper presents MambaBEV, a novel BEV-based 3D object detection model that integrates Mamba2 into a camera-based perception framework. 
We design the TemporalMamba block to effectively fuse temporal information and enhance global context awareness.
Extensive experiments on the nuScenes dataset demonstrate the effectiveness and efficiency of the proposed approach, particularly in improving large-object detection accuracy. Furthermore, evaluation within an end-to-end autonomous driving framework confirms its strong performance in motion forecasting and planning tasks.
This work highlights the potential of state-space models for autonomous driving perception and provides a promising direction for improving global context modeling in BEV-based systems.
\vspace{-6mm}

\balance
\bibliographystyle{unsrt} 
\bibliography{reference}

\end{document}